# Application of CARE-SD text classifier tools to assess distribution of stigmatizing and doubt-marking language features in EHR

**Drew Walker, PhD[1], Jennifer Love, MD, MSCR[2], Swati Rajwal, MTech[3], Isabel C Walker, PA-C[4], Hannah LF Cooper, PhD[5], Abeed Sarker, PhD[6,7], Melvin Livingston III, PhD[5]**

**[1]Department of Health Systems Science, Kaiser Permanente School of Medicine, Pasadena, CA, USA ; [2]Department of Emergency Medicine, Mount Sinai, New York, NY, USA; [3]Department of Computer Science, Emory College of Arts and Sciences, Emory University, Atlanta, GA, USA; [4]Children's Heart Center, Children's Healthcare of Atlanta, Atlanta GA, USA; [5]Department of Behavioral, Social, Health Education Sciences, Rollins School of Public Health, Emory University, Atlanta GA, USA; [6]Department of Biomedical Informatics, School of Medicine, Emory University, Atlanta GA, USA ; [7]Department of Biomedical Engineering, Georgia Institute of Technology and Emory University, Atlanta, GA, USA**

**Abstract**

*Introduction: Electronic health records (EHR) are a critical medium through which patient stigmatization is perpetuated among healthcare teams. Methods: We identified linguistic features of doubt markers and stigmatizing labels in MIMIC-III EHR via expanded lexicon matching and supervised learning classifiers. Predictors of rates of linguistic features were assessed using Poisson regression models. Results: We found higher rates of stigmatizing labels per chart among patients who were Black or African American (RR: 1.16), patients with Medicare/Medicaid or government-run insurance (RR: 2.46), self-pay (RR: 2.12), and patients with a variety of stigmatizing disease and mental health conditions. Patterns among doubt markers were similar, though male patients had higher rates of doubt markers (RR: 1.25). We found increased stigmatizing labels used by nurses (RR: 1.40), and social workers (RR: 2.25), with similar patterns of doubt markers. Discussion: Stigmatizing language occurred at higher rates among historically stigmatized patients, perpetuated by multiple provider types.*

**Introduction**

Electronic health record (EHR) notes are important mediums in which providers may use stigmatizing language and stereotyping in ways that work to further enforce and create multiple forms of structural marginalization.[1] Recent research in linguistic bias and stigmatization by Beukeboom and Burgers has produced the Social Categories and Stereotypes Communication Framework, which can be applied in efforts to understand and investigate provider language in healthcare.[2] This framework posits that stereotypes and stigmas exist in a shared cognitive space among groups, shaped by communicative context and content. Several forms of stigmatizing language within provider notes have been identified, focused on the expression of degree of trust or doubt in patient testimony through use of stigmatizing labels , as well as "doubt markers".[3–5] Stigmatizing labels to describe groups are used to perpetuate stereotypes, exaggerate essentialist group differences, and when used by providers, can lead to reduced trust and communication among patients. A recent NIDA publication described a list of words to avoid using around patients with substance use disorders, including "addict", "abuser", "user", or "junkie", which have been found to be associated with perceived stigmatization by patients.[6] Similar studies have been applied to other chronic illness populations, identifying terms like "sickler" or "frequent flier" which may be used to further stigmatize patients with chronic illnesses who are often admitted into the hospital.[4,6,7] Doubt markers, or evidentials, which are defined as "the linguistic coding of epistemology",[8] are words that are frequently used in chart language to question the veracity of patients. Among the many words used as evidentials, words and expressions used to confer doubt or uncertainty such as: allegedly, apparently, or verbs like claimed, are often used to when describing patient testimonies, for example: "patient claimed their pain was 10/10".[3] Qualitative research on the transmission of provider bias through clinical notes has identified common manifestations of both negative and positive bias within patient charts,[9] and preliminary research into this text and language-based transmission of bias towards patients has found associations with disparities in care across race, gender, drug use, mental health disorders, and other marginalized conditions.[1,3,9–11]

As qualitative research on stigmatizing provider language has grown, natural language processing techniques are only beginning to be applied to allow up for stigmatizing language on large-scale EHR databases. To date, most NLP studies on stigmatizing language have involved solely regular expression matching among smaller samples of patient notes.[12] Existing studies have pointed to the need to incorporate data-driven lexicon expansion and refined supervised classification approaches to tackle the classification of such nuanced and continuously evolving topics as provider stigma.[13,14] This study applies the CARE-SD: classifier-based analysis for recognizing provider stigmatizing and doubt marker labels in EHR, a tool comprised of both data-driven lexicon expansion and supervised learning approaches on a large scale, widely accessible dataset.[15] The methods and design of this study allow our results to be easily built upon by other researchers for further validation, benchmarking, and model improvement. Following EHR text classification, we examine the distribution of stigmatizing and doubt-marking language features across patient- and provider-level predictors.

## Methods

### *MIMIC-III Dataset*

The Medical Information Mart for Intensive Care, or "MIMIC-III", is a database of comprehensive, de-identified EHR, free-text notes, and event documentation for over 40,000 patients admitted to the ICU at Beth Israel Deaconess Medical Center in Boston, MA from 2001 to 2012.[16] This dataset contains over 1.2 million clinical provider notes, across nearly 50,000 admissions. IRB access to this dataset was completed by research team through MIT and PhysioNet's Data Use Agreement and CITI Data research training credentialing. [16]

### *Data preparation*

In the data preparation phase of our study, we employed a comprehensive approach to process and analyze the textual data from the MIMIC-III dataset, described in greater detail in the attached Supplemental File. Sentence-level classifications of doubt-markers and stigmatizing labels were aggregated as outcomes, derived from the CARE-SD analytic toolkit.

### *CARE-SD: Classifier-based Analysis for Recognizing and Eliminating Stigmatizing and Doubt Marker Labels in Electronic Health Records*

Lexicons and supervised learning classifiers for stigmatizing labels and doubt markers were derived from the "CARE-SD: classifier-based analysis for recognizing provider stigmatizing and doubt marker labels in electronic health records" toolkit.[15] This toolkit combines lexicon-based matching with supervised and semi-supervised learning classifiers to identify stigmatizing language based on human annotation of clinical notes.

### *Stigmatizing labels*

Stigmatizing label lexicon development for was guided by literature on stigmatizing language in medical care, specifically from the NIDA "Words Matter" publication, Sun's "Negative Patient Descriptors: Documenting Racial Bias in the Electronic Health Record", and Zestcott's "Health Care Providers' Negative Implicit Attitudes and Stereotypes of American Indians".[6,17,18] The initial stem word list consisted of 18 words: "abuser","junkie","alcoholic", "drunk", "drug-seeking","nonadherent", "agitated", "angry", "combative", "noncompliant", "confront", "noncooperative", "defensive", "hysterical", "unpleasant", "refuse","frequent-flier", "reluctant". This list was subsequently expanded using contextual word embeddings models and GPT 3.5 models and pruned by expert clinical and public health annotators. Supervised learning natural language processing classifiers were developed from an annotated sample derived from MIMIC-III charts with matching sentences containing lexicon terms, resulting in a logistic regression classifier model with 81% accuracy, 75% precision, 84% recall, and .79 macro-F1 score, comparable to human annotator agreement, (87%, kappa = .74).

### *Doubt markers*

Doubt marker lexicon development was guided by literature on use of "doubt markers" in medical care, specifically led by Beach and colleagues, which identified words such as "claims", "insists", and "adamant" or "apparently", which have been found to be used to discredit or invalidate patient testimony.[3] The 6 words included on the initial stem list were: "adamant", "claimed", "insists", "allegedly","disbelieves","dubious".

The CARE-SD original validation article provides a final lexicon list used to develop annotation samples used to refine the supervised learning models, for both stigmatizing labels and doubt markers.[15] The doubt marker sentence classifier used is a RoBERTa model with 86% accuracy, 86% precision, 71% recall, and .84 macro-F1 score, also comparable to human annotator agreement (87%, kappa = .73). Model training code for this study is available via GitHub.[19]

### *Patient/Provider Clustering*

We calculated median incidence rate ratios using multilevel Poisson models to compare the clustering of stigmatizing labels and doubt markers within notes at the patient and provider levels. This measure helps to describe the relative change in the rate of stigmatizing labels and doubt markers per chart, when comparing identical charts from two randomly selected different rate-ordered clusters.[20] A value of 1 suggests completely independent note samples, with larger values indicating greater proportion of linguistic feature rates per note variance explained by a clusters at the patient and provider levels.

***Poisson Regressions assessing patient/provider predictors on stigmatizing labels and doubt markers***
We assessed differences in the rates of stigmatizing labels and doubt marker linguistic features per chart, per patient and provider, using Poisson Regression models, with patient and provider-level variables in separate models. In each model, we included the offset of the natural log of the total number of charts to account for differences in frequencies in charts across patients and providers. The model form for both the patient-level and provider-level models are provided below.

*Model Set 1 and 2: Patient and provider-level predictors of stigmatizing labels, doubt markers*
$\log(\lambda_{\text{pt or provider}}$ [Stigmatizing Labels or Doubt Markers]) = $\beta 0+\beta 1(\text{Predictor})X_{\text{pt or provider}} + \log(\text{Total \# of Charts})$

Patient-level predictors included Race/ethnicity, which was recategorized by AW to account for low cell size, insurance type (where Government insurance represents non-Medicare and non-Medicaid types of government-supported insurance, including programs from the Department of Defense TRICARE, Veterans Health Administration program, or Indian Health Service), gender, age category aligned with Medline MeSH Age Categories,[21] as well as whether or not patients had ICD-9 codes associated with Sickle Cell Disease, Opioid Use Disorder, Obesity, Symptomatic HIV, Substance Use Disorder, Schizophrenia, Mood Disorder, Anxiety, PTSD, Suicide Atempts, and Suicidal Ideation. Provider-level predictors included provider type, which was categorized from MIMIC-III free-text fields by AW and IW. Providers with unknown category and pharmacist category were excluded from regressions.

**Results**
Table 1 offers descriptive statistics on patient demographics, provider types, and distributions of stigmatizing labels and doubt markers within charts.

**Table 1.** Demographic descriptive statistics on patients, provider types, and distributions of stigmatizing labels and doubt markers per charts in MIMIC-III dataset

| | **Overall (N=11630 Patients)** |
|---|---|
| **Race/Ethnicity*** | |
| White | 8312 (71.5%) |
| Asian | 353 (3.0%) |
| Black/African American | 945 (8.1%) |
| Hispanic/Latino | 440 (3.8%) |
| Native American/Alaskan Native | 16 (0.1%) |
| Other | 345 (3.0%) |
| Unknown/Declined | 1219 (10.5%) |
| **Age: Mean (SD)** | 62.6 years (16.6) [15.2, 89.0] |
| **Insurance** | |
| Private | 4509 (38.8%) |
| Government^ | 353 (3.0%) |
| Medicaid | 1211 (10.4%) |
| Medicare | 5450 (46.9%) |
| Self Pay | 107 (0.9%) |
| **Gender** | |
| Female | 5086 (43.7%) |
| Male | 6544 (56.3%) |
| **Diagnoses** | |
| Sickle Cell Disease | 25 (0.2%) |
| Opioid Use Disorder | 217 (1.9%) |
| Obesity | 674 (5.8%) |
| HIV | 115 (1.0%) |
| Substance Use Disorder | 1355 (11.6%) |
| Schizophrenia | 32 (0.3%) |
| Mood Disorder | 701 (6.0%) |
| Anxiety | 492 (4.2%) |
| PTSD | 67 (0.6%) |
| Suicide Attempts | 56 (0.5%) |
| Suicidal Ideation | 20 (0.2%) |
| **Stigmatizing Labels Count Per Patient** | |
| Mean (SD) | .5 (1.99) |
| Median [Min, Max] | 0 [0, 90] |
| **Doubt Marker Labels Count Per Patient** | |
| Mean (SD) | 0.09 (0.48) |
| Median [Min, Max] | 0 [0, 29] |
| **Provider Types*** | (N = 1880 Providers) |
| APPs (NPs, PA-Cs) | 31(1.6%) |
| Pharmacists | 4 (0.2%) |
| Physicians | 590 (31.4%) |
| Registered Dieticians | 23 (1.2%) |
| Registered Nurses | 1004 (53.4%) |
| Rehab (OTs/PTs) | 49 (2.6%) |
| Respiratory Therapists | 42 (2.2%) |
| Social Workers | 47 (2.5%) |
| Unknown** | 90 (4.8%) |

Most patients were White (71.5%) and Male (56.3%), and Medicare (46.9%) was the most common insurance type. Registered nurses (53.4%) and physicians (31.4%) composed most of this sample. Across clusters, both stigmatizing labels and doubt markers were found to have non-normal right-skew distributions. Spearman rank correlation test showed significant moderate correlation between doubt marker classification labels and stigmatizing classification labels at the patient level (Rho = .1887, p <.001), with higher correlation at the provider level (Rho = .4459, p < .001). Clustering of stigmatizing labels (Median IRR: 7.08) and doubt markers (Median IRR: 147.8) was highest at the patient level. Patient-level models results are displayed in Table 2.

**Table 2.** Poisson regression results showing relationships between demographic patient predictor variables associations with stigmatizing linguistic EHR features per chart. Rate Ratios (95%CI)

| | **Stigmatizing Classifier Labels** | **Doubt Marker Classifier Labels** |
|---|---|---|
| **Gender (Ref = Female)** | 1.02 (.97, 1.08) | **1.25 (1.11, 1.42)**** |
| **Ethnicity (Ref = White)** | | |
| Asian | **0.54 (0.45, 0.65)**** | **0.46 (0.27, 0.72)**** |
| Black/African American | **1.16 (1.08, 1.25)**** | 1.06 (0.88, 1.27) |
| Hispanic/Latino | **0.74 (0.63, 0.86)**** | 0.78 (0.53, 1.10) |
| Native American/Alaskan Native | **0.24 (0.08, 0.56)*** | 1.01 (0.25, 2.63) |
| Other | **0.42 (0.34, 0.52)**** | **0.49 (0.30, 0.75)*** |
| Unknown/Declined | **0.66 (0.59, 0.73)**** | **0.75 (0.59, 0.93)*** |
| **Insurance (Ref= Private)** | | |
| Government-run^ | **2.46 (2.32, 2.61)**** | **2.32 (2.03, 2.67)**** |
| Self-Pay | **2.12 (1.45, 2.95)**** | **4.94 (2.69, 8.25)**** |
| **Diagnoses** | | |
| HIV (Symptomatic) | **2.59 (2.16, 3.08)**** | **2.40 (1.51, 3.60)**** |
| Obesity | **1.98 (1.83, 2.14)**** | **2.14 (1.78, 2.55)**** |
| Opioid Use Disorder | **2.79 (2.48, 3.11)**** | **3.80 (2.98, 4.77)**** |
| Sickle Cell Disease | 0.41 (0.10, 1.07) | 1.54 (0.26, 4.75) |
| Substance Use Disorder | **1.87 (1.74, 2.01)**** | **2.38 (2.02, 2.78)**** |
| Schizophrenia | 1.40 (.67, 2.53) | 1.73 (.29, 5.35) |
| Mood Disorder | **2.11 (1.93, 2.32)**** | **1.50 (1.15, 1.91)*** |
| Anxiety | **1.67 (1.48, 1.87)**** | **2.06 (1.59, 2.62)**** |
| PTSD | **3.82 (3.07, 4.69)**** | **2.92 (1.56, 4.92)**** |
| Suicide Attempts | **2.77 (2.18, 3.45)**** | **4.34 (2.73, 6.50)**** |
| Suicidal Ideation | **4.90 (2.76, 7.94)**** | 3.89 (.65, 12.04) |
| **Age (Ref = Middle Aged (45-64)** | | |
| Adolescent (13-18) | 1.19 (.54, 2.23) | 0.83 (0.05, 3.67) |
| Adult (19-44) | **1.32 (1.22, 1.43)**** | **1.28 (1.05, 1.55)*** |
| Aged (65-79) | **0.90 (.84, .96)*** | 0.87 (.74, 1.02) |
| Aged, 80 and over (>80) | 1.03 (0.95 1.12) | **0.79 (.63, .97)*** |

***p is significant at <.05 value**
****p is significant at <.0001 value**
**^ Government-run includes the MIMIC-III insurance categories of "Government", "Medicare", and "Medicaid"**

Male patients had 1.25 times the rate of doubt markers per chart than females (95% CI: 1.11, 1.42, p <.0001). Compared to White patients, Black or African American patients received 1.16 higher rates of stigmatizing labels per chart. Compared to patients with private insurance, patients with Government-run insurance (encompassing all government employee care, Medicaid, or Medicare) had 2.46 times as many Stigmatizing Labels per chart and 2.32 times as many Doubt Markers. Similarly, patients with no insurance, labeled "Self Pay", had 2.12 times as many Stigmatizing Labels per chart, and 4.94 as many Doubt Markers per chart as patients with Private insurance.

Patients with symptomatic HIV had 2.59 times the rate of Stigmatizing Labels per chart, and 2.40 times the rate of Doubt Markers per chart, compared with patients without HIV. Patients with obesity had 1.98 times higher rates of Stigmatizing labels per chart, and 2.14 times higher rates of doubt markers per chart. Patients with opioid use disorder experienced some of the highest differences in rates of both stigmatizing labels per chart (2.79 times higher) and doubt markers per chart (3.80 times higher). Comparatively, substance use disorder was associated with 1.87 times higher rates of stigmatizing labels, and 2.28 times higher rates of doubt markers per chart. Nearly all mental health conditions were associated with higher rates of both stigmatizing labels and doubt markers. Patients diagnosed with PTSD also 3.82 times the rates of stigmatizing labels per chart and 2.92 times the rates of doubt markers per chart. Suicidal attempts and ideation were also associated with higher rates of both stigmatizing labels and doubt markers. Patients with Sickle Cell Disease and schizophrenia had no significant difference in either type of stigmatizing linguistic feature compared to patients without these conditions.

Out of all age categories, adults ages 19-44 had the highest rates of stigmatizing labels and doubt markers, receiving 1.32 times rates of doubt markers as compared with Middle aged adults, the reference group comprising the largest segment of the MIMIC-III patient sample, and 1.28 times higher rates of doubt markers. Provider type regression model results are displayed in Table 3.

**Table 3.** Poisson regression results, modeled at the provider level, showing relationships between provider types with stigmatizing linguistic EHR features per chart. Rate Ratios (95%CI)

| | **Stigmatizing Classifier Labels** | **Doubt Marker Classifier Labels** |
|---|---|---|
| **Provider Type (Ref = Physicians)** | | |
| Advanced Practice Providers (NPP, PA-C) | **0.15 (0.10, 0.21)**** | **0.04 (0.00, 0.17)**** |
| Registered Dieticians | **0.25 (0.15, 0.37)**** | 1.18 (0.63, 2.00) |
| Registered Nurses | **1.40 (1.30, 1.50)**** | **1.46 (1.21, 1.77)**** |
| Rehab (OT/PT) | 1.18 (0.86, 1.58) | **2.27 (1.19, 3.92)*** |
| Respiratory Therapists | 0.92 (0.65, 1.26) | 0.35 (0.06, 1.09) |
| Social Workers | **2.25 (1.76, 2.84)**** | **5.27 (3.33, 7.98)**** |

***p is significant at <.05 value**
****p is significant at <.0001 value**
**Pharmacists removed from regression analyses due to low cell size (n = 4)**

Compared with Physicians, several groups of providers were found to have significantly different rates of stigmatizing label and doubt markers per chart. Advanced Practice Providers (including Physician Associates and Nurse Practitioners) had 0.15 times the rates of Stigmatizing labels per chart, and only .04 times the rates of Doubt Marker labels per chart. Registered Dieticians were also found to have .25 times the rate of stigmatizing labels per chart as Physicians. Registered Nurses had 1.40 times higher rates of Stigmatizing labels and 1.46 times higher rates of doubt marker labels than physicians. Rehab care team members, consisting of Occupational and Physical Health Therapists, were found to have 2.27 times the rate of doubt markers as Physicians. Finally, Social Workers reported the highest rates of stigmatizing labels and doubt markers, using 2.25 times the rate of stigmatizing labels per chart, and 5.27 times the rate of doubt markers per chart.

## Discussion and Conclusions

Results from this study largely support understanding of stereotyping and stigmatization in language provided by the SCSC Framework. Our analyses found that the distribution of stigmatizing labels and doubt markers is largely a collective process, with systematic differences observed among patients with historically marginalized identities based on gender, race, insurance, and illness.

Median incident rate ratios across both stigmatizing labels and doubt markers indicated drastically higher rates of clustering by patient- than by provider-level. This supports the SCSC central assertion that stereotypes are a result of shared cognitive and group-level processes.[2] Results from correlations between doubt markers and stigmatizing labels indicate that these features are frequently used together and may in tandem work to further stigmatize or invalidate patients' testimonies. Changing language in medical notes requires us to understand the role of ingroup dynamics among a medical team, and how it may be "relationally beneficial" to perpetuate a stereotype to foster similarity and agreeableness.[22] This social pressure may contribute to the increased effort it takes to refute a collectively held stereotype.[23,24] Further, the processes of note segment duplication and replication may make it even more difficult to refute stereotypes and change language around patients once they begin to be described a certain way.[25] This highlights the need for systems in place such as the ones developed and tested within this study to identify stigmatizing language in real time.

The differences in distributions of stigmatizing labels and doubt markers within electronic health records largely fit the pattern of demonstrating disproportionately higher rates across historically stigmatized and marginalized groups. Our finding of 25% higher rates of doubt markers among notes of male patients compared to female patients stands in contrast to previous work, which identified higher rates of doubt marking language and pain disbelief among female patients [3,26] While the current study stands to illuminate trends on a broad level, our doubt marker classification tools can be used to draw more targeted samples to guide further qualitative inquiry to identify the specific clinical contexts in which doubt marking language may be differentially applied by gender. There may be certain diagnoses or clinical situations in this ICU setting, such as alcohol or substance-related conditions, which occur more frequently among men and may also coincide with higher rates of doubt marker usage.[27] Black/African American patients received an estimated 1.16 times higher rates of stigmatizing classifier labels per note, when compared to White patients. This finding is in line with previous work which has identified disproportionately higher rates of stigmatizing or doubt-marking language in notes of patients who are Black or African American, as well as broader trends of provider-held implicit biases.[28,29]

Insurance, as a proxy indicator for socioeconomic status, revealed higher rates of stigmatizing labels among patients receiving government assistance (Medicare, Medicaid) for insurance, and especially for patients with no insurance, identified as “Self-Pay”, when compared with patients with private insurance. This increase is greater in the latter group, which could be conceptualized as a group with the lowest socioeconomic status. [30–32]

Of the frequently stigmatized diagnoses selected for this study, disproportionately higher rates of doubt markers and stigmatizing labels were found for patients with obesity, symptomatic HIV, and were highest among patients with opioid use disorder. These trends were largely congruent with a wealth of research documenting experiences of stigmatization, ridicule, and testimonial injustice across these groups.[3,33,34] An absence of difference among patients with sickle cell disease and schizophrenia may also be driven by an overall low sample size of patients with these conditions within this MIMIC-III dataset sample.

While rates of stigmatizing labels and doubt markers identified among patients with mental health diagnoses also aligns with patient experiences of stigmatization,[35] these differences in language features may also be driven by fears of legal or regulatory issues, prompting providers to create additional evidential distance to specify uncertainty around patient attitudes, behaviors, or testimonies.[36]

Nurses used stigmatizing labels and doubt markers at 1.40 and 1.46 times higher rates than physicians. Because nurses spend more time with patients, they frequently endure greater frequencies of negative behaviors in forms of verbal or physical abuses from patients and families that may arise out of patient frustrations or conditions.[37] It is possible that these frustrations may be reflected in increased use of stigmatizing labels and doubt markers used among nurses. As nurses occupy a lower power status within the hierarchy of medical teams, they may also have higher motivation to generate in-group bonding among the medical team, support beliefs in agreement with stereotypes held by higher-status team members, and face greater risks with going against shared team stereotypes and assumptions about patients.[38–40] Social workers had the highest rates of both stigmatizing and doubt markers usage within notes. As ancillary health professionals within the healthcare team chain of command, social workers may face similar pressures to agree or conform to group norms as nurses.

While the results of this paper have illuminated important trends in the distribution of stigmatizing linguistic features in the EHR, it is not without limitations. Firstly, due to the predominantly White sample of MIMIC-III, our race and ethnicities had to be simplified to incorporate these factors into analyses. This simplification likely results in some degree of erasure or potentially grouping of ethnicities and races which may have very different histories of stigmatization within the medical setting. Analyses in this study also did not incorporate time into analyses. Further work incorporating how time affects the accumulation of stigmatizing and doubt-marking linguistic features is needed to see how patient reputations may emerge and change over time.

This study successfully demonstrated the utility of stigmatizing language detection models to assess differences in how patients from a variety of backgrounds and diagnoses are discussed by providers within the EHR. Trends identified in this study should underscore the strong and present threat of stigmatizing and doubt marking language that exists disproportionately among highly marginalized patient groups. CARE-SD tools can be applied to assess the distribution of linguistic stigmatizing and doubt-marking features across a wide variety of other EHR settings, populations, and care outcomes. Large language models, which were implemented in the CARE-SD lexicon development stage, could also be incorporated through pretraining the models to query more nuanced questions about patient stigmatization patterns, summarize common situations in which stigmatizing language and doubt markers are used, or destigmatize medical language once identified.

## References

1. FitzGerald C, Hurst S. Implicit bias in healthcare professionals: a systematic review. BMC Med Ethics 2017;18(1):19; doi: 10.1186/s12910-017-0179-8.
2. Beukeboom CJ, Burgers C, Reviews R of CR-O-AH-QL. How Stereotypes Are Shared Through Language - A Review and Introduction of the Social Categories and Stereotypes Communication (SCSC) Framework. Rev Commun Res 2019.
3. Beach MC, Saha S, Park J, et al. Testimonial Injustice: Linguistic Bias in the Medical Records of Black Patients and Women. J Gen Intern Med 2021; doi: 10.1007/s11606-021-06682-z.
4. P Goddu A, O'Conor KJ, Lanzkron S, et al. Do Words Matter? Stigmatizing Language and the Transmission of Bias in the Medical Record. J Gen Intern Med 2018;33(5):685–691; doi: 10.1007/s11606-017-4289-2.
5. Zhang H, Lu AX, Abdalla M, et al. Hurtful Words: Quantifying Biases in Clinical Contextual Word Embeddings. In: Proceedings of the ACM Conference on Health, Inference, and Learning. CHIL '20 Association for Computing Machinery: New York, NY, USA; 2020; pp. 110–120; doi: 10.1145/3368555.3384448.
6. Abuse NI on D. Words Matter - Terms to Use and Avoid When Talking About Addiction. 2021. Available from: https://www.drugabuse.gov/nidamed-medical-health-professionals/health-professions-education/words-matter-terms-to-use-avoid-when-talking-about-addiction [Last accessed: 1/22/2022].
7. Glassberg J, Tanabe P, Richardson L, et al. Among emergency physicians, use of the term "Sickler" is associated with negative attitudes toward people with sickle cell disease. Am J Hematol 2013;88(6):532–533; doi: 10.1002/ajh.23441.
8. Chafe WL, Nichols J. Evidentiality: The Linguistic Coding of Epistemology. Ablex; 1986.
9. Park J, Saha S, Chee B, et al. Physician Use of Stigmatizing Language in Patient Medical Records. JAMA Netw Open 2021;4(7):e2117052–e2117052; doi: 10.1001/jamanetworkopen.2021.17052.
10. Ashford RD, Brown AM, Curtis B. Substance use, recovery, and linguistics: The impact of word choice on explicit and implicit bias. Drug Alcohol Depend 2018;189:131–138; doi: 10.1016/j.drugalcdep.2018.05.005.
11. Volkow ND, Gordon JA, Koob GF. Choosing appropriate language to reduce the stigma around mental illness and substance use disorders. Neuropsychopharmacology 2021;46(13):2230–2232; doi: 10.1038/s41386-021-01069-4.
12. Kelly PJA, Snyder AM, Agénor M, et al. A scoping review of methodological approaches to detect bias in the electronic health record. Stigma Health 2023;No Pagination Specified-No Pagination Specified; doi: 10.1037/sah0000497.
13. Harrigian K, Zirikly A, Chee B, et al. Characterization of Stigmatizing Language in Medical Records. In: Proceedings of the 61st Annual Meeting of the Association for Computational Linguistics (Volume 2: Short Papers). (Rogers A, Boyd-Graber J, Okazaki N. eds) Association for Computational Linguistics: Toronto, Canada; 2023; pp. 312–329; doi: 10.18653/v1/2023.acl-short.28.
14. Barcelona V, Scharp D, Moen H, et al. Using Natural Language Processing to Identify Stigmatizing Language in Labor and Birth Clinical Notes. Matern Child Health J 2024;28(3):578–586; doi: 10.1007/s10995-023-03857-4.
15. Walker A, Thorne A, Das S, et al. CARE-SD: classifier-based analysis for recognizing provider stigmatizing and doubt marker labels in electronic health records: model development and validation. J Am Med Inform Assoc 2024;ocae310; doi: 10.1093/jamia/ocae310.
16. Johnson AEW, Pollard TJ, Shen L, et al. MIMIC-III, a freely accessible critical care database. Sci Data 2016;3(1):160035; doi: 10.1038/sdata.2016.35.
17. Sun M, Oliwa T, Peek ME, et al. Negative Patient Descriptors: Documenting Racial Bias In The Electronic Health Record. Health Aff (Millwood) 2022;41(2):203–211; doi: 10.1377/hlthaff.2021.01423.
18. Zestcott CA, Spece L, McDermott D, et al. Health Care Providers' Negative Implicit Attitudes and Stereotypes of American Indians. J Racial Ethn Health Disparities 2021;8(1):230–236; doi: 10.1007/s40615-020-00776-w.
19. Walker D. Drew-Walkerr/CARE-SD-Stigma-and-Doubt-EHR-Detection. 2024.
20. Austin PC, Stryhn H, Leckie G, et al. Measures of clustering and heterogeneity in multilevel Poisson regression analyses of rates/count data. Stat Med 2018;37(4):572–589; doi: 10.1002/sim.7532.
21. Kastner M, Wilczynski NL, Walker-Dilks C, et al. Age-Specific Search Strategies for Medline. J Med Internet Res 2006;8(4):e25; doi: 10.2196/jmir.8.4.e25.

22. Clark AE, Kashima Y. Stereotypes help people connect with others in the community: A situated functional analysis of the stereotype consistency bias in communication. J Pers Soc Psychol 2007;93(6):1028–1039; doi: 10.1037/0022-3514.93.6.1028.
23. Bratanova B, Kashima Y. The “Saying Is Repeating” Effect: Dyadic Communication Can Generate Cultural Stereotypes. J Soc Psychol 2014;154(2):155–174; doi: 10.1080/00224545.2013.874326.
24. Bargh JA. The Cognitive Monster: The Case against the Controllability of Automatic Stereotype Effects. In: Dual-Process Theories in Social Psychology The Guilford Press: New York, NY, US; 1999; pp. 361–382.
25. Steinkamp J, Kantrowitz JJ, Airan-Javia S. Prevalence and Sources of Duplicate Information in the Electronic Medical Record. JAMA Netw Open 2022;5(9):e2233348; doi: 10.1001/jamanetworkopen.2022.33348.
26. Hoffmann DE, Tarzian AJ. The girl who cried pain: a bias against women in the treatment of pain. J Law Med Ethics J Am Soc Law Med Ethics 2001;29(1):13–27; doi: 10.1111/j.1748-720x.2001.tb00037.x.
27. Westerhausen D, Perkins AJ, Conley J, et al. Burden of Substance Abuse-Related Admissions to the Medical ICU. Chest 2020;157(1):61–66; doi: 10.1016/j.chest.2019.08.2180.
28. Gopal DP, Chetty U, O’Donnell P, et al. Implicit bias in healthcare: clinical practice, research and decision making. Future Healthc J 2021;8(1):40–48; doi: 10.7861/fhj.2020-0233.
29. Hoffman KM, Trawalter S, Axt JR, et al. Racial bias in pain assessment and treatment recommendations, and false beliefs about biological differences between blacks and whites. Proc Natl Acad Sci U S A 2016;113(16):4296–4301; doi: 10.1073/pnas.1516047113.
30. Brown CE, Engelberg RA, Sharma R, et al. Race/Ethnicity, Socioeconomic Status, and Healthcare Intensity at the End of Life. J Palliat Med 2018;21(9):1308–1316; doi: 10.1089/jpm.2018.0011.
31. Monuteaux MC, Du M, Neuman MI. Evaluation of Insurance Type as a Proxy for Socioeconomic Status in the Pediatric Emergency Department: A Pilot Study. Ann Emerg Med 2024; doi: 10.1016/j.annemergmed.2023.12.013.
32. Jang B-S, Chang JH. Socioeconomic status and survival outcomes in elderly cancer patients: A national health insurance service-elderly sample cohort study. Cancer Med 2019;8(7):3604–3613; doi: 10.1002/cam4.2231.
33. Richard P, Ferguson C, Lara AS, et al. Disparities in physician-patient communication by obesity status. Inq J Med Care Organ Provis Financ 2014;51:0046958014557012; doi: 10.1177/0046958014557012.
34. Geter A, Herron AR, Sutton MY. HIV-Related Stigma by Healthcare Providers in the United States: A Systematic Review. AIDS Patient Care STDs 2018;32(10):418–424; doi: 10.1089/apc.2018.0114.
35. Vistorte AOR, Ribeiro WS, Jaen D, et al. Stigmatizing attitudes of primary care professionals towards people with mental disorders: A systematic review. Int J Psychiatry Med 2018;53(4):317–338; doi: 10.1177/0091217418778620.
36. Balfour ME, Hahn Stephenson A, Delany-Brumsey A, et al. Cops, Clinicians, or Both? Collaborative Approaches to Responding to Behavioral Health Emergencies. Psychiatr Serv 2022;73(6):658–669; doi: 10.1176/appi.ps.202000721.
37. Edward K, Stephenson J, Ousey K, et al. A systematic review and meta-analysis of factors that relate to aggression perpetrated against nurses by patients/relatives or staff. J Clin Nurs 2016;25(3–4):289–299; doi: 10.1111/jocn.13019.
38. Noyes AL. Navigating the Hierarchy: Communicating Power Relationships in Collaborative Health Care Groups. Manag Commun Q 2022;36(1):62–91; doi: 10.1177/08933189211025737.
39. Corley MC, Goren S. The Dark Side of Nursing: Impact of Stigmatizing Responses on Patients. Sch Inq Nurs Pract 1998;12(2):99-110,112-118.
40. Adams LY, Maykut CA. Bullying: The Antithesis of Caring Acknowledging the Dark Side of the Nursing Profession. Int J Caring Sci 2015;8(3):765–773.

## Supplemental File Section 1: Data Preparation

Our pre-processing steps to prepare the MIMIC-III dataset, included removing all duplicate charts, as well as those labeled as EEG or Radiology, in order to restrict to charts more likely to have subjective narrative and patient history text data.

Following this, we ran previously developed supervised learning classifier models, trained by clinical annotators, to identify and classify these language features to match instances in which human clinical annotators deemed language to be stigmatizing our doubt-inducing more precisely. These supervised learning models were applied on any sentences which contained lexicon matches from the stigmatizing labels and doubt marker word list.

We then merged the predictive sentence-level labels with all sentences in the MIMIC-III dataset. Next, we then aggregated the presence of stigmatizing labels and doubt markers at the note level. We then merged caregiver data from the caregiver table using the Caregiver Identifier (CGID), and patient data from the patients table using the Subject Identifier (SubjectID). Patient insurance and ethnicity were merged from the Admissions table (HADM_ID), and in order to simplify analyses, we selected the first appearing insurance and ethnicity value listed in the dataset for each patient. Patient race/ethnicity free-text categories were organized into a smaller number of distinct categories to facilitate use in regression analyses. We also re-organized the free-text provider type label fields into distinct categories, which was guided by Society of Critical Care Medicine provider type categories, and conducted by IW, who has clinical intensive care unit experience as a PA-C. [52] Provider type and race/ethnicity labels and categories are provided in the following section. Patient diagnoses were derived as binary variables (having the diagnosis or not) by linking ICD-9 patient codes from the "diagnoses_icd" table, using regular expressions to capture relevant codes for sickle cell disease , opioid use disorder (OUD), HIV (symptomatic), and obesity.

To facilitate analysis at both the patient and provider levels, we summarized outcomes of stigmatizing labels and doubt markers at these respective levels. Prior to model building, we conducted univariate and bivariate descriptive analyses to assess distribution of outcomes and predictor variables across levels of patients and providers. We also assessed bivariate correlations of the patient and provider level outcomes of stigmatizing labels and doubt markers.

## Supplemental File Section 2: Patient Ethnicity and Provider Recategorizations

### Provider recategorization

| provider_label_final | Unique_Provider_Labels |
|---|---|
| Physicians | MDs, Res, MD, MSIV, Med St, RO, HMS MS, DML, MDS, Mds, MedRes, HMSIV, MSV, HMS IV, RF, MS V, MedSt, md |
| APP | PA, NP, SNP, NNP, SNNP, nnp |
| Pharmacist | PharmD, RPh, RPH |
| Registered Dieticians | RD, DI, RD,LDN, MS,RD, RD/LDN |
| Registered Nurses | RN, CoOpSt, RRT, SN, rn, RNs, NSV, RNC, Rn, CoOPSt, StNurs, CoOpst, CCRN, RNBA, CoWker, DRM, SRN, Nurs, NS, StNRS, Nurse, CM, StNur, RNStu, NStude, CoWkr, StuNur, Co-Wkr, PracSt, CO-Op |
| Rehab (OT/PT) | Rehab, OTR/L, IMD, PT |
| Respiratory Therapist | SRT, RT, RRts, RTStu, CRT, RRTs, RTS, RRt, rrt |
| Social Workers | SW, LICSW, CRS, MSWint, SWInt, SW Int |

# Supplemental File Section 3: Lexicons for Doubt Markers and Stigmatizing Labels

| Lexicon | Stem Word List | Expanded Words (Pruned to) | GPT-3.5 added words | High-noise terms removed | Final Lexicon | Final Lexicon Length |
|---|---|---|---|---|---|---|
| **Doubt Markers** | "adamant", "claimed", "insists", "allegedly","disbelieves","dubious" | 60 total, reduced to 42. Agreement = 80% | [' "skeptical', ' dubiousness', ' questionable', ' doubting', ' uncertain', ' skepticalness', ' incredulous', ' hesitating', ' suspicious', ' mistrustful', ' distrustful', | 'suspicion', 'suspicious', 'questionable', 'questioning', 'uncertain', 'hesitancy', 'hesitant','unsure' | ['"doubtful', '"dubious", '.insists', 'accused', 'adamant', 'adamant/belligerant', 'adamantly', 'addamant', 'alledgedly', 'alleged', 'allegedly', 'allegedly-unnecessary', 'asserted', 'believes', 'claimed', 'claimedthat', 'claimes', 'claiming', 'confessionally', | 58 |
| | | | ' unconvinced',<br>' unsure',<br>' hesitant',<br>' wary',<br>' dubious',<br>' disbelieving',<br>' skepticalism',<br>' hesitancy',<br>' skepticism',<br>' mistrust',<br>' uncertainness',<br>' disbelief',<br>' suspicion',<br>' mistrustfulness',<br>' incredulity',<br>' incredulously',<br>' wavering',<br>' ambivalent',<br>' waveringly',<br>' questionableness',<br>' mistrustingly',<br>' doubter',<br>' questioning',<br>' doubtingly',<br>' mistrusting',<br>' doubtful',<br>' skeptic',<br>' unconvincedly',<br>' mistrustingly',<br>' mistrustfully',<br>' doubtingness',<br>' skepticism',<br>' questioningness',<br>' unbelieving',<br>' unsureness',<br>' skepticness',<br>' questioningness',<br>' doubtingly',<br>' unbelievingly',<br>' skeptically',<br>' mistrustingly',<br>' mistrustfully',<br>' skeptically', | | 'culpably', 'disbelief', 'disbelieve', 'disbelieved', 'disbeliever', 'disbelievers', 'disbelieves', 'disbelieving', 'disclaimed', 'doggedly', 'doubious', 'doubtful', 'dubious', 'dubious/equivocal', 'dubiously', 'insisist', 'insisisted', 'insist', 'insisted', 'insisting', 'insists', 'misbelieve', 'misbelieved', 'misbelieves', 'mistrustful', 'mistrusting', 'non-dubious', 'proclaimed', 'purportedly', 'reinsists', 'skeptical', 'speculative', 'supposedly', 'them-insists', 'unconvinced', 'undisguisedly', 'unreliable', 'unsure'", 'wavering'] | |
| | | | ' questioningly' , ' doubtingly', ' skeptically', ' mistrustingly', ' mistrustfully'" ] | | | |
| **Stigmatizing Labels** | "abuser","junkie","alcoholic", "drunk", "drug-seeking","nonadherent", "agitated", "angry", "combative", "noncompliant", "confront", "noncooperative", "defensive", "hysterical", "unpleasant", "refuse","frequent-flyer", "reluctant" | 180 , reduced to 83. Annotator agreement = .75. | [' "hysterical', ' aggressive', ' drug addict', ' non-compliant', ' lazy', ' attention-seeking', ' manipulative', ' hypochondriac', ' difficult', ' mentally unstable', ' troublemaker', ' irresponsible', ' unpredictable' , ' irrational', ' needy', ' demanding', ' disruptive', ' uncooperative ', ' unreliable', ' high maintenance', ' attention-seeker', ' dramatic', ' attention-seeking', ' lazy', ' invalid', ' faker', ' irrational', ' hostile', ' aggressive', ' challenging', ' uncooperative ', ' deceptive', ' demanding', ' unreliable', ' high-strung', ' self-destructive', ' unstable', ' manipulative', ' entitled', ' attention-seeking', ' violent', ' drug seeker', ' malingerer', ' faker', ' mentally ill', ' dangerous', ' | 'difficult', 'suspicious','aggressive','unstable', 'dramatic', 'unreliable','entitled','invalid',' violent', 'dangerous' | ['"hysterical', '"drug-seeking", '"hysterical", '"drug-seeking", '"junkie", '.reluctant', 'abuse/abuser', 'abused-abuser', 'abuser', "abuser's", 'abusers', 'addictive-drug-seeking', 'alcoholic', 'angry', 'angry-disgusted', 'angry/disgusted', 'attention-seeker', 'attention-seeking', 'challenging', 'combative', 'combatively', 'compliant/noncompliant', 'counterdefensive', 'deceptive', 'defensive', 'defensive/offensive', 'delusional', 'demanding', 'disruptive', 'drug addict', 'drug seeker', 'drug-craving/drug-seeking', 'drug-seeking', 'drug-seeking/-taking', 'drug-seeking/drug-taking', 'drug-seeking/taking', 'drug-seeking/use', 'drunk', 'drunken', 'drunkenly', 'drunker', 'drunkest', 'drunks', 'ex-abuser', 'ex-alcoholic', 'faker', 'frequent-flier', 'frequent-flyer', 'frequent-flyers', 'frequent-fvl', 'frequent-hitter', 'frequent-hitters', 'high maintenance', 'high-strung', 'histrionic-hysterical', 'hostile', 'hypochondriac', 'hypochondriac-hysterical', 'hysteric', 'hysterical', 'hysterical-obsessive', | 127 |
| | | | delusional', ' needy', ' overly sensitive', ' unstable', ' irrational'"] | | 'hysterical/anaclitic', 'hystericals', 'hysterics', 'incompliant', 'irrational', 'irrational'", 'irresponsible', 'iv-abuser', 'ivdabuser', 'junkie', "junkie's", 'junkies', 'lazy', 'ma-abuser', 'malingerer', 'manipulative', 'mentally ill', 'mentally unstable', 'morereluctant', 'needy', 'non-adherent', 'non-alcoholic/alcoholic', 'non-compliant', 'non-cooperating', 'non-cooperation', 'non-cooperative', 'non-cooperatively', 'nonadhered', 'nonadherent', 'nonadherently', 'nonadherents', 'noncompliant', 'noncompliant/compliant', 'noncompliant\\medically', 'noncompliants', 'noncooperating', 'noncooperation', 'noncooperative', 'noncooperatively', 'novelty/drug-seeking', 'ny-nonadherent', 'onadherent', 'overdefensive', 'overly sensitive', 'prealcoholic', 'pt.noncompliant', 'refuse', 'refuses', 'refusing', 'reluctanly', 'reluctant', 'reluctantly', 'reluctants', 'schizo-hysterical', 'self-destructive', 'troublemaker', 'un-adherent', 'unadherent', 'uncooperative', 'unpleasant', 'unpleasant/annoying', 'unpleasantly', 'unpleasantries', 'unpredictable', 'unwilling', 'unwillingly'] | |